\documentclass[conference]{IEEEtran}
\IEEEoverridecommandlockouts
\usepackage{cite}
\usepackage{amsmath,amssymb,amsfonts}
\usepackage{algorithmic}
\usepackage{graphicx}
\usepackage{textcomp}
\usepackage{xcolor}
\usepackage{makecell}
\usepackage{upgreek}
\usepackage{float}
\usepackage{tabularx}
\usepackage{booktabs}
\usepackage{hyperref}

\def\BibTeX{{\rm B\kern-.05em{\sc i\kern-.025em b}\kern-.08em
    T\kern-.1667em\lower.7ex\hbox{E}\kern-.125emX}}
\begin{document}

\title{FuseMamba-VD: Dual Branch VideoMamba with Gated Class Token Fusion for Violence Detection}

\author{Damith Chamalke Senadeera$^{1,2}$, Awais Rauf$^{3}$, Shibo Li$^{1, 2}$,  Dimitrios Kollias$^{1,2}$, Gregory Slabaugh$^{1,2}$\\
$^{1}$Queen Mary University of London, UK\\
$^{2}$Digital Environment Research Institute (DERI), London, UK\\
$^{3}$Queen's University Belfast, UK\\
{\tt\small d.c.senadeera@qmul.ac.uk, a.rauf@qub.ac.uk, \{shibo.li, d.kollias, g.slabaugh\}@qmul.ac.uk}
}

\maketitle

\begin{abstract}
The rapid proliferation of surveillance cameras has increased the demand for automated violence detection. While CNNs and Transformers have shown success in extracting spatio-temporal features, they struggle with long-term dependencies and computational efficiency. We propose FuseMamba-VD: Dual Branch VideoMamba with Gated Class Token Fusion (GCTF), an efficient architecture combining a dual-branch design and a state-space model (SSM) backbone where one branch captures spatial features, while the other focuses on temporal dynamics. The model performs continuous fusion via a gating mechanism from the spatial branch into the temporal branch to enhance detection of violent activities even in challenging surveillance scenarios. We also present a new benchmark by merging RWF-2000, RLVS, SURV and VioPeru datasets in video violence detection, ensuring strict separation between training and testing sets. Experimental results demonstrate that our model achieves state-of-the-art performance on this benchmark and also on DVD dataset which is a recently introduced dataset on video violence detection, offering an optimal balance between accuracy and computational efficiency, demonstrating the promise of SSMs for scalable, resource efficient video violence detection. The code and pre-trained models are available at \url{https://github.com/damith92/FuseMamba-VD}.
\end{abstract}

\section{Introduction}
\label{sec:intro}

The widespread deployment of low-cost surveillance cameras in public and private spaces has led to an explosion of video data. Continuous human monitoring of these feeds is neither practical nor reliable for detecting violent behavior, motivating robust and efficient deep learning approaches for automated surveillance violence detection.

\begin{figure}[t]
  \centering
   \includegraphics[width=1\linewidth]{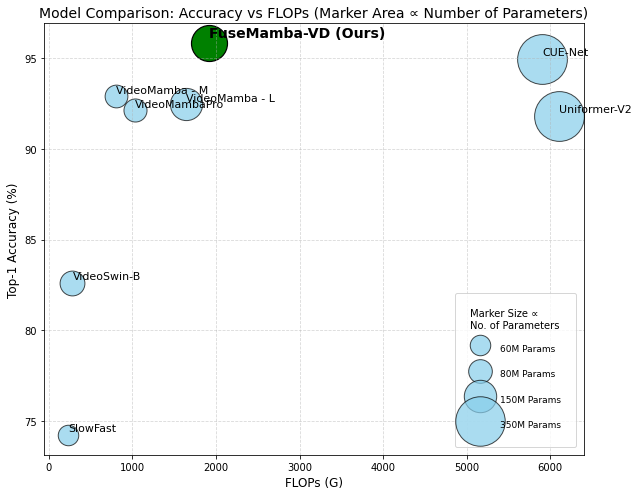}
   \vspace{-10pt}
   \caption{Model comparison on the combined dataset showing Top-1 Accuracy vs.\ FLOPs, with marker size proportional to the number of parameters.}
   \vspace{-10pt}
   \label{fig:5}
\end{figure}

Recent work has leveraged 3D convolutional neural networks (CNNs) and Transformer-based architectures to extract spatio-temporal features from videos~\cite{action_recognition_VD_1}. While 3D CNNs capture local motion patterns, they struggle with long-range dependencies, whereas Transformers model global context but incur quadratic complexity with sequence length, limiting scalability~\cite{uniformerv2}. State-space models (SSMs) offer an attractive alternative, providing linear-time sequence processing with strong modeling capacity~\cite{mamba_main}.

In this work, we propose \emph{FuseMamba-VD}, an SSM-based architecture tailored for supervised violence detection. We design a dual-branch architecture with a spatial-first branch that emphasizes fine-grained spatial cues and a temporal-first branch that focuses on motion dynamics. The two branches interact via a continuous gated fusion of their CLS tokens across Mamba layers, enabling progressive refinement of spatial and temporal representations and improving sensitivity to subtle motion differences indicative of violence.

To enhance robustness and generalizability, we also construct an amalgamated surveillance benchmark by combining four strongly labeled open-source violence datasets: Real World Fighting (RWF-2000)~\cite{rwf_dataset}, Real Life Violence Situations (RLVS)~\cite{reallife_violence_dataset}, SURV~\cite{surv_dataset}, and VioPeru~\cite{vioperu_1}, while carefully preventing data leakage between training and testing splits. This integrated dataset spans diverse scenes, camera viewpoints, and environmental conditions.

Our contributions are threefold:
\begin{enumerate}
    \item \textbf{Dual-branch state-space architecture:} We introduce the first state-space-based dual-branch design for supervised video violence detection, explicitly decoupling spatial and temporal reasoning while enabling continuous semantic interaction via class tokens, in contrast to prior dual-encoder SSM or CNN+Transformer models that rely on late fusion or heavy attention mechanisms.
    \item \textbf{Gated Class Token Fusion (GCTF):} We propose a continuous, layer-wise fusion mechanism that adaptively integrates branch-specific class tokens across Mamba layers, allowing gradual alignment of spatial and temporal features without overcommitting to either branch early in the network.
    \item \textbf{Integrated surveillance benchmark and efficiency:} We curate an integrated surveillance violence dataset from RWF-2000, RLVS, SURV, and VioPeru with strict data leakage control between training and testing sets, and demonstrate that \emph{FuseMamba-VD} achieves state-of-the-art performance on both this combined dataset and the DVD dataset~\cite{dvd_dataset} while substantially reducing parameters and FLOPs compared to prior work (Fig.~\ref{fig:5}, Tab.~\ref{tab:model_comparison}), making it a promising solution for real-world surveillance applications.
\end{enumerate}

\section{Related Work}
\label{sec:related_work}

Research on automated violence detection largely follows two paradigms: anomaly detection and action recognition. In anomaly detection, violent events are treated as rare deviations from normal behavior~\cite{anaomaly_VD_1}, which can work in controlled settings but often fails to capture the complex context of real-world surveillance scenes~\cite{ucfcrime_dataset}. In contrast, action-recognition-based methods formulate violence detection as supervised classification, using 3D CNNs~\cite{action_recognition_VD_1} and skeleton based models to model human interactions~\cite{pose_VD}, with more recent work evaluating on real surveillance datasets such as RWF-2000~\cite{rwf_dataset} instead of earlier non-surveillance benchmarks like Hockey Fight~\cite{hockeyfight}. Transformer-based and hybrid CNN–Transformer models~\cite{swintransformer_VD, cue-net, uniformerv2} further leverage self-attention to capture global context while retaining local modeling via convolutions, but their quadratic attention complexity can limit scalability on long, high-resolution surveillance videos.

Recent progress in subquadratic state-space model (SSM) architectures~\cite{mamba_main} has enabled more efficient video understanding. VideoMamba~\cite{videomamba_1}, replaces self-attention with a linear-complexity SSM module that preserves long-range dependency modeling while improving scalability for video.  VideoMambaPro~\cite{vmambapro} which is the latest extension of VideoMamba, further extends this line by introducing masked backward computation and elemental residual connections to address the decay of long-range information and conflicts between token representation issues identified in VideoMamba. A key trade-off, however, is the added architectural complexity relative to vanilla VideoMamba, since these improvements rely on extra design mechanisms, even though the reported parameter count and FLOPs remain comparable. To the best of our knowledge, our paper is the first to investigate the use of a dual branch SSM for video violence detection in the supervised learning space.

\section{Proposed Method}
\label{sec:methedology}

Our design utilizes efficient bidirectional state-space modules to introduce a novel dual branch architecture. We employ two parallel pipelines, each implementing a variant of VideoMamba with distinct scanning strategies, with the intention of separately extracting spatial and temporal features from video inputs, combined with a cropping module to focus on human-interactions. We introduce a novel Gated Class Token Fusion (GCTF) mechanism that combines information between the two branches. GCTF is performed at each layer in the network, providing a form of continuous fusion. The enriched feature representation is combined from the Spatial-First Scanning Branch to the Temporal-First Scanning Branch to form a unified representation for final classification.

\subsection{Overview of Mamba Architectures}

The Mamba family of architectures rethinks the standard self-attention mechanism by leveraging efficient state-space models (SSMs) for long-range dependency modeling ~\cite{mamba_main}. The key idea is to view a sequence as the output of a continuous dynamical system, where the evolution of a hidden state is governed by ordinary differential equations~\cite{mamba_main}. In its continuous form, a state-space model is given by:
\begin{equation}
    h'(t) = \mathbf{A} h(t) + \mathbf{B} x(t), \quad y(t) = \mathbf{C} h(t),
    \label{eq:continuous_ssm_expanded}
\end{equation}
where $x(t)$ is the input at time $t$, $h(t) \in \mathbb{R}^{N}$ is the hidden state, and where $\mathbf{A} \in \mathbb{R}^{N \times N}$ represents the evolutionary matrix of the system and $\mathbf{B} \in \mathbb{R}^{N \times D}$, and $\mathbf{C} \in \mathbb{R}^{d \times N}$ are projection matrices. To process discrete token sequences, this continuous system is approximated via discretization (commonly using a zero-order hold), which includes a timescale parameter ${\mathbf \Updelta}$ to transform the continuous parameters ${\mathbf A}, {\mathbf B}$ to their discrete learnable counterparts $\overline{{\mathbf A}}, \overline{{\mathbf B}}$:
\begin{equation}
    \overline{{\mathbf A}} = \exp({\mathbf \Updelta \mathbf A}), \overline{{\mathbf B}} = ({\mathbf \Updelta \mathbf A})^{-1} (\exp({\mathbf \Updelta \mathbf A}) - {\mathbf I}) \cdot {\mathbf \Updelta \mathbf B}
    \label{eq:discrete_ssm_expanded1}
\end{equation}
\begin{equation}
    h_t = \overline{\mathbf A} h_{t-1} + \overline{ \mathbf B} x_t, \quad y_t = \mathbf{C} h_t.
    \label{eq:discrete_ssm_expanded2}
\end{equation}
Contrary to traditional models that primarily rely on linear time-invariant SSMs, Mamba~\cite{mamba_main} implements a Selective Scan Mechanism (S6) as its core SSM operator. Within S6, the parameters ${\mathbf B} \in \mathbb{R}^{B\times L \times N}$, ${\mathbf C} \in \mathbb{R}^{B\times L \times N}$, and ${\mathbf \Updelta} \in \mathbb{R}^{B \times L \times D}$ are directly derived from the input data $x \in \mathbb{R}^{B \times L \times D}$, demonstrating an inherent aptitude for capturing context by  dynamically adjusting weights.

\subsection{FuseMamba-VD}

\begin{figure*}
  \centering
   \includegraphics[width=0.85\linewidth]{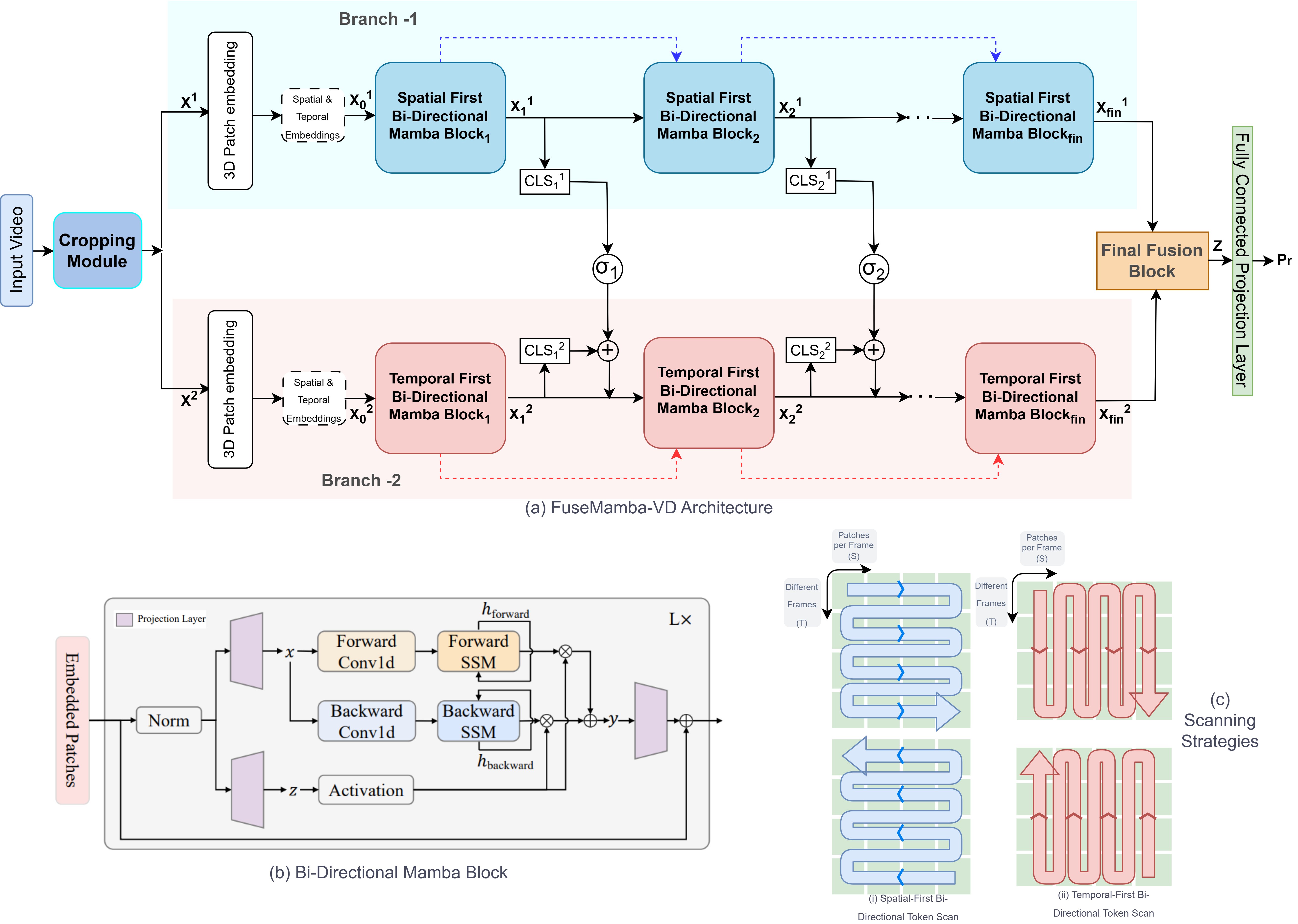}
   \vspace{-8pt}
   \caption{ Part (a): The overall FuseMamba-VD Architecture with: 1) Cropping module to detect people and crop spatially the area of focus; 2) Branch-1 and Branch-2 which implements VideoMamba architecture in two parallel pathways with lateral connections between parallel layers implementing continuous Gated Class Token Fusion (GCTF); 3) Final Fusion Block. Part (b): Bi-Directional Vision Mamba Encoder Block used in the above architecture. Part (c): (i) Spatial-First Scanning. (ii) Temporal-First Scanning. }
   \vspace{-12pt}
  \label{fig:1}
\end{figure*}

We introduce a novel architecture, FuseMamba-VD for Violence Detection in surveillance videos as shown in Fig.~\ref{fig:1}.  The architecture contains four main components, namely: (a) Cropping Module; (b) Branch-1 (Spatial-First Scanning); (c) Branch-2 (Temporal-First Scanning); and (d) Final Fusion Block; inspired by the motivational factors discussed in the preceding section.

\subsubsection{Cropping Module}

To focus on human actions, our cropping mechanism extracts the region that encompasses all detected people in each frame to guide the network, based on the observation that violent incidents typically involve multiple individuals. This approach enhances the model’s ability to accurately identify violent behavior as evident in the ablation study.

\subsubsection{Dual-Branch Architecture Overview}

Let $\mathbf{X} \in \mathbb{R}^{3 \times T \times H \times W }$ denote an input video, where $T$ is the number of frames, $H \times W$ represents the spatial resolution, and $3$ stands for RGB channels. Separately in each of the two branches, the video is first tokenized into patch embeddings where 3D convolution (\textit{i.e.}, 1$\times$16$\times$16) is used to project the input videos $\mathbf{X}^{v}\in \mathbb{R}^{3\times T\times H\times W}$ into $L$ non-overlapping spatiotemporal patches $\mathbf{X}^{p}\in \mathbb{R}^{L\times C}$,
where $L$$=$$t$$\times$$h$$\times$$w$ ($t$$=$$T$, $h$$=$$\frac{H}{16}$, and $w$$=$$\frac{W}{16}$).
Then the sequence of tokens will be padded with a  $\mathbf{X}_{cls}$ learnable class token along with positional embeddings $\mathbf{p}_{s} \in \mathbb{R}^{(hw+1) \times C}$ and temporal embeddings $\mathbf{p}_{t} \in \mathbb{R}^{t \times C}$.
\begin{align}
    \mathbf{X}^{1} ={}& \left[ \mathbf{X}_{cls}^{1}, \mathbf{X}^{1} \right] + \mathbf{p}_{s}^{1} + \mathbf{p}_{t}^{1},
\end{align}
\begin{align}
    \mathbf{X}^{2} ={}& \left[ \mathbf{X}_{cls}^{2}, \mathbf{X}^{2} \right] + \mathbf{p}_{s}^{2} + \mathbf{p}_{t}^{2},
\end{align} 
and subsequently are fed into the two distinct pipelines.

\paragraph{Branch-1 (Spatial-First Scanning Pipeline)} This branch reorganizes the tokens such that the spatial layout within each frame is prioritized. Tokens are grouped and ordered by their spatial coordinates before being concatenated across frames as depicted in Fig.~\ref{fig:1} (c)(i). This strategy is aimed at learning fine-grained local spatial features that are critical for recognizing visual cues. This Spatial-First scan in this branch is performed bi-directionally.

\paragraph{Branch-2 (Temporal-First Scanning Pipeline)} Here, the tokens are reorganized based on the frame and then stacked along the spatial dimension maintaining their natural temporal order so that the sequential progression of frames is preserved as depicted in Fig.~\ref{fig:1} (c)(ii). This ordering can facilitate the model to capture dynamic motion patterns and global temporal dependencies pertaining to each of those tokens, which are essential for detecting violent actions and changes over time. Similar to Branch-1, this scan is also performed bi-directionally. 

This strategy enables leveraging the pre-trained weights of a vanilla VideoMamba to initialize each branch in our architecture. Pre-trained initialization is beneficial because it provides a robust starting point, accelerates convergence during training, and enhances overall performance by transferring learned representations from extensive prior training~\cite{uniformerv2}. During the forward pass, Branch-1 processes its input video to produce a sequence of features, including a dedicated class (CLS$^{1}$) token for each layer. Meanwhile, Branch-2 receives two inputs: the video embedding tokens padded with the CLS$^{2}$ token and additional intermediate (CLS$^{1}$) token features from each block of Branch-1 (excluding the final block). This second input allows Branch-2 to incorporate spatial context learned from Branch-1 into its feature learning to produce better results as evident by the results in Tab.~\ref{tab:fusion_mechanisms} of the ablation study. To  enhance information flow, residual skip connections are also incorporated among the blocks within each branch.

\subsubsection{Gated Class Token Fusion (GCTF) between Branch-1 and Branch-2}

After processing each block in the network, both pipelines output their respective CLS tokens:

\begin{itemize}
    \item $\text{CLS}_{l}^{1} \in \mathbb{R}^{d}$ from the spatial-first scanning Branch-1,
    \item $\text{CLS}_{l}^{2} \in \mathbb{R}^{d}$ from the temporal-first scanning Branch-2.
\end{itemize}

To combine these complementary features, a learnable gate is applied between each parallel block. The gating mechanism is defined as:
\begin{equation}
    \sigma'_{l}={}Sigmoid(\sigma_{l})
    \label{eq:gate_equation}
\end{equation}
\begin{equation}
    \text{CLS}_{\text{fused(}l\text{)}}^{2} =  \sigma'_{l} \odot \text{CLS}_{l}^{2} + \left(1 -  \sigma'_{l}\right) \odot \text{CLS}_{l}^{1},
    \label{eq:gate_fusion}
\end{equation}
where $\sigma_{l} \in \mathbb{R}^{d}$ is a learnable parameter vector for block $l$, and it is passed through the Sigmoid function to ensure that the gate value lies between 0 and 1 before performing element-wise multiplication denoted by $\odot$ with CLS tokens from each branch to dynamically weigh the best contributions. Unlike typical fusion schemes that operate at a single depth or combine all tokens, GCTF selectively merges class-level semantics across branches at every depth, enabling layer-wise refinement and preventing early overcommitment to either modality.

\subsection{Final Fusion Block} 

At the very end of the architecture, a final fusion block integrates the final CLS tokens from Branch-1 and Branch-2 through concatenation. The output $\mathbf{Z} \in \mathbb{R}^{2d}$ is obtained as:
\begin{equation}
   \mathbf{Z} = \operatorname{concat}(\text{CLS}_{fin}^{1}, \text{CLS}_{fin}^{2}),
    \label{eq:final_z}
\end{equation}

Finally, the target class $Pr$ is obtained by passing $\textbf{Z}$ through a fully connected projection layer.

\section{Experiments and Results}
\label{sec:exp}

\subsection{Datasets and Implementation Details}

\begin{table*}[t]
\centering
\caption{Comparative results among various architectures for the newly combined dataset and the DVD dataset. Each model is characterized by its backbone type (CNN, Transformer, SSM), the pretraining dataset, the input resolution, the number of parameters (in millions), the FLOPs (measured for the full pipeline, including any auxiliary modules), the Top-1 Accuracy (\%), and the F1-scores (\%) for Violent and Non-Violent classes.}
\vspace{-4pt}
\setlength{\tabcolsep}{5pt}
\renewcommand{\arraystretch}{1.0}
\resizebox{\textwidth}{!}{
\begin{tabular}{|l|p{2.6cm}|l|l|c|c|c|c|c|c|c|c|}
\hline
\textbf{Architecture} & \makecell*[c]{\textbf{Model}} & \textbf{Pretraining} & \textbf{Input Size} &
\textbf{Params. (M)} & \textbf{FLOPs (G)} &
\multicolumn{3}{c|}{\textbf{Combined Dataset (\%)}} & \multicolumn{3}{c|}{\textbf{DVD Dataset (\%)}} \\
\cline{7-9} \cline{10-12}
 & & & & & & \textbf{Top-1} & \textbf{F1-V} & \textbf{F1-NV} &
                   \textbf{Top-1} & \textbf{F1-V} & \textbf{F1-NV} \\
\hline
CNN            & SlowFast          & -      & 64$\times$224$^{2}$ & 60    & 234   & 74.21 & 70.46 & 77.21 & 61.72 & 55.85 & 66.22 \\
\hline
Trans.         & VideoSwin-B       & K-400  & 64$\times$224$^{2}$ & 88    & 281.6 & 82.62 & 83.49 & 81.85 & 64.30 & 60.40 & 67.49 \\
\hline
CNN+Trans.     & Uniformer-V2      & K-400  & 64$\times$224$^{2}$ & 354   & 6108  & 91.81 & 91.93 & 91.68 & 70.95 & 63.36 & 75.94 \\
\hline
CNN+Trans.     & CUE-Net           & K-400  & 64$\times$224$^{2}$ & 354   & 5906  & 94.97 & 94.92 & 95.02 & 73.68 & 68.71 & 77.28 \\
\hline
SSM            & VideoMamba-M        & K-400  & 64$\times$224$^{2}$ & 74    & 806   & 92.90 & 92.80 & 92.99 & 72.47 & 66.42 & 76.67 \\
\hline
SSM            & VideoMambaPro-M        & K-400  & 64$\times$224$^{2}$ & 76    & 1028   & 92.13 & 90.75 & 93.16 & 72.01 & 68.05 & 75.10 \\
\hline
SSM            & VideoMamba-L        & K-400  & 64$\times$224$^{2}$ & 148    & 1644   & 92.46 & 92.22 & 92.68 & 71.10 & 63.76 & 75.97 \\
\hline
SSM            & \makecell*[l]{\textbf{FuseMamba-VD (ours)}} & K-400  & 64$\times$224$^{2}$ & 154.3 & 1912  & \textbf{95.85} & \textbf{95.89} & \textbf{95.81} & \textbf{74.13} & \textbf{68.97} & \textbf{77.82} \\
\hline
\end{tabular}
}
\vspace{-12pt}
\label{tab:model_comparison}
\end{table*}

\subsubsection{Combined Dataset of RWF-2000, RLVS, SURV, and VioPeru}

We construct a unified real-world violence benchmark by merging four strongly labeled datasets: Real-World Fighting (RWF-2000)~\cite{rwf_dataset}, Real Life Violence Situations (RLVS)~\cite{reallife_violence_dataset}, SURV~\cite{surv_dataset}, and VioPeru~\cite{vioperu_1}. RWF-2000, SURV, and VioPeru contain exclusively surveillance footage, while RLVS additionally includes unconstrained online videos. All clips are short (2--5\,s) and depict real-world scenarios, in contrast to staged or acted datasets such as AUTO~\cite{auto_dataset}. We also exclude Movie Scene Violence~\cite{penet2012multimodal} and Hockey Fight~\cite{hockeyfight}, which are sourced purely from films and sports broadcasts, as they are less representative of real surveillance conditions. We further exclude weakly labeled datasets such as UCF-Crime~\cite{ucfcrime_dataset} and XD-Violence~\cite{xdviolence}, whose video-level annotations (Violent / Non-Violent) do not indicate when violent actions occur, making them incompatible with our strongly labeled setup. To avoid cross-dataset leakage, we first respect each dataset’s official train/test split and then run a duplicate check using VideoMAE~\cite{video_mae} embeddings and cosine similarity across all train/test videos. Pairs with similarity $\geq 0.75$ are manually inspected, and one duplicated video is removed from the RLVS test split. The final combined dataset contains 3{,}664 training clips (1{,}832 violent / 1{,}832 non-violent) and 915 testing clips (457 violent / 458 non-violent).

\subsubsection{DVD Dataset}

DVD~\cite{dvd_dataset} is a recent large-scale dataset for violence detection, consisting of 344 long videos (2.7M frames) with frame-level violence annotations covering diverse environments, lighting conditions, camera viewpoints, and social interactions. To use DVD in a clip-based setting, we follow the official frame-level labels and segment each video into contiguous violent and non-violent clips, grouping consecutive frames with the same label into a single clip.

This produces 2{,}648 clips in total: 1{,}099 violent ($\approx 9.5$ hours) and 1{,}549 non-violent ($\approx 15$ hours), amounting to $\approx 24.6$ hours of footage. To avoid data leakage, all clips originating from the same source video are assigned exclusively to either the training or test split. We adopt an approximate 80/20 split that preserves the violent/non-violent balance, yielding 1{,}987 training clips (820 violent, 1{,}167 non-violent) and 661 test clips (279 violent, 382 non-violent).

\subsubsection{Implementation Details}

Our algorithm was implemented with the specifications of the VideoMamba-Middle (M) architecture with 32 layers where the hidden dimension $d$ was 576 across the layers. Following the standard practice in~\cite{videomamba_1}, we applied uniform frame sampling across each video. Moreover, all the baselines were trained with the same pre-processing pipeline. Also, all FLOPs reported in this paper correspond to the end-to-end pipeline, i.e., they include any auxiliary modules required such as cropping module, etc.

\subsection{Results}

We evaluate all models on the combined dataset and the DVD dataset using Top-1 accuracy and class-wise F1 scores, following prior work on these benchmarks~\cite{cue-net, swintransformer_VD, rwf_dataset, vioperu_1}. Tab.~\ref{tab:model_comparison} summarizes the results. On the combined dataset, our FuseMamba-VD consistently outperforms all baselines. While SlowFast~\cite{slowfast_1} (which is a dual-branch 3D CNN architecture), VideoSwin~\cite{videoswintrf}, UniFormer-V2~\cite{uniformerv2}, the VideoMamba \& VideoMambaPro variants~\cite{videomamba_1, vmambapro}, and CUE-Net~\cite{cue-net} already achieve strong performance (up to 94\% accuracy), our model attains 95.85\% accuracy with F1 scores of 95.89\% and 95.81\% for the violent and non-violent classes, respectively.

The DVD dataset is markedly more challenging due to its real-world variability, class imbalance (279 violent vs.\ 382 non-violent clips) and clip length variability. Here, our FuseMamba-VD again delivers the best performance, reaching 74.13\% accuracy with F1 scores of 68.97\% (violent) and 77.82\% (non-violent), surpassing SlowFast, VideoSwin, UniFormer-V2, VideoMamba \& VideoMambaPro variants, and CUE-Net (all $\leq 71\%$ Top-1 accuracy). The higher F1 score for the non-violent class reflects the class imbalance, but the improved violent-class F1 demonstrates that FuseMamba-VD remains more reliable in detecting the minority class, which is also the more practically critical class with violent events here.

These gains are achieved not merely by increasing model size, but with a relatively compact model through more effective feature modeling and fusion. FuseMamba-VD has 154\,M parameters and 1912\,GFLOPs, more than halving both parameters and FLOPs compared to CUE-Net (354\,M and 5906\,GFLOPs). Since CUE-Net is the closest competitor to FuseMamba-VD in accuracy, we also compare inference latency on an NVIDIA H100 GPU where CUE-Net achieves 72.64 FPS, while FuseMamba-VD reaches 343.53 FPS, giving an $\approx$4.7$\times$ speed-up. Against the similarly sized VideoMamba-Large (148\,M parameters), our model improves accuracy from 92.46\% to 95.85\% on the combined dataset and from 71.10\% to 74.13\% on DVD. McNemar’s two-sided exact test~\cite{mcnemar} confirms that these improvements on similar sized SSM models are statistically significant on both datasets (all $p < 0.05$).

\subsection{Visual Analysis of Results}

\begin{figure}
  \centering
   \includegraphics[width=1\linewidth]{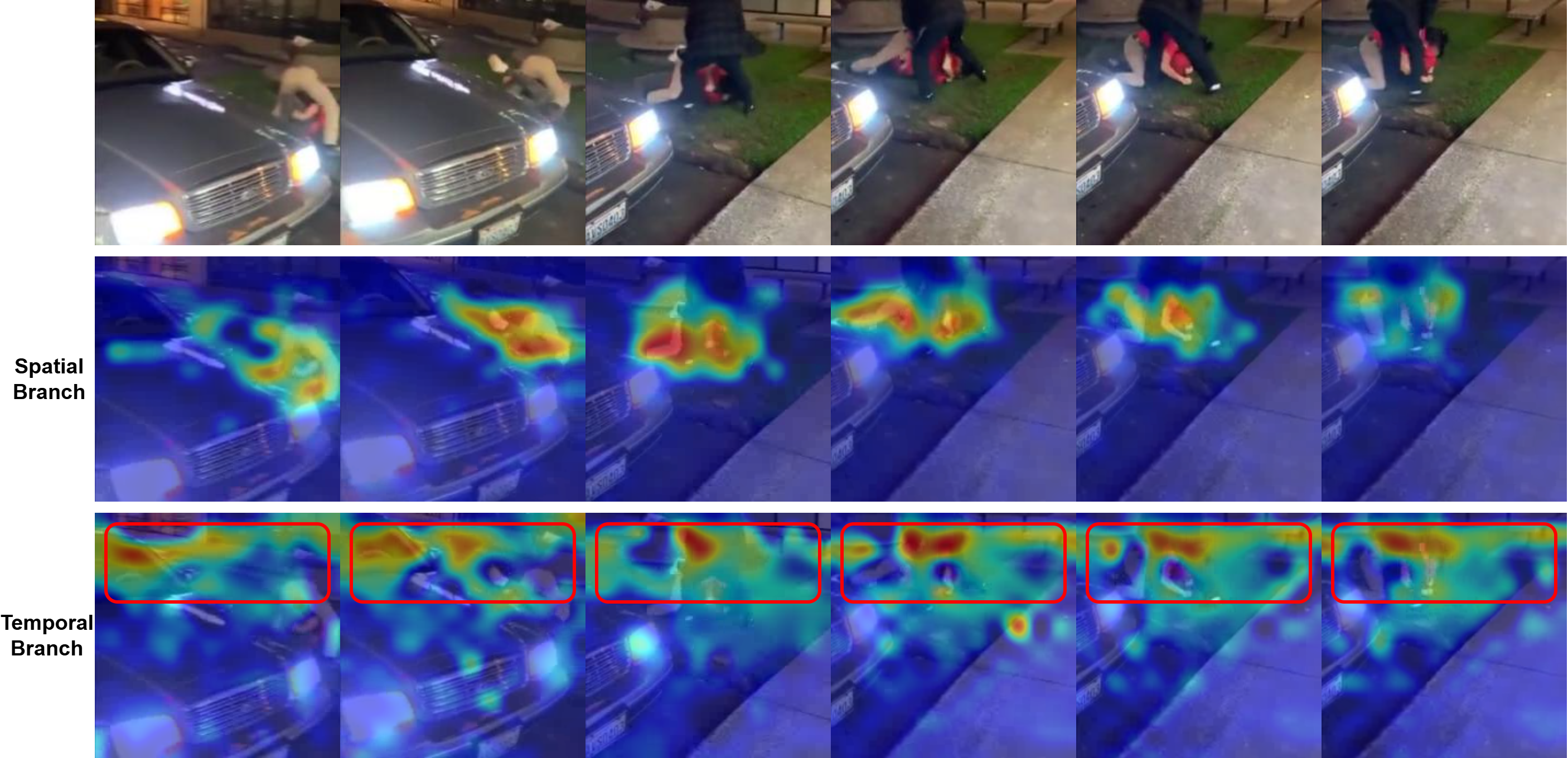}
   \vspace{-14pt}
   \caption{Visual Analysis of Class Activation Maps (CAMs). CAMs for a violent-labeled video correctly classified as violent, illustrating that the model accurately focuses on regions of human interaction, even under occlusion and near frame edges.}
   \vspace{-13pt}
  \label{fig:3}
\end{figure}

We employ Grad-CAM and Grad-CAM++~\cite{gradcamplusplus} to visualize the regions influencing the model’s predictions. In the spatial-first branch, Grad-CAM cleanly highlights discriminative regions captured by the Mamba blocks, whereas the temporal-first branch focuses on motion dynamics, compressing spatial detail and spreading evidence over multiple locations and frames. Because this multi-region evidence makes standard Grad-CAM maps diffuse, we use Grad-CAM++ on the temporal branch, which better captures the spread of motion cues. In Fig.~\ref{fig:3}, the spatial branch sharply localizes a partially occluded violent interaction, while the temporal branch yields a more distributed pattern that traces the incident’s trajectory across frames, effectively learning where in the frame violence is likely to emerge at some point in the video.

\subsection{Ablation Study}

\begin{table}[!h]
\centering
\caption{Performance comparison on accuracy with and without the cropping module and the residual skip connections among the blocks in each branch.}
\vspace{-4pt}
\setlength{\tabcolsep}{5pt}
\renewcommand{\arraystretch}{1.0}
\resizebox{\columnwidth}{!}{%
\begin{tabular}{l|cc|cc}
\hline
\textbf{Model} & \multicolumn{2}{c|}{\textbf{Cropping - $\times$}} & \multicolumn{2}{c}{\textbf{Cropping - $\checkmark$}} \\
\cline{2-5}
 & \textbf{Skip - $\times$} & \textbf{Skip - $\checkmark$ } & \textbf{Skip - $\times$} & \textbf{Skip - $\checkmark$} \\
\hline
VideoMamba - M (Spatial First) & 92.90\% & 94.00\% & 93.44\% & 94.54\% \\
VideoMamba - L (Spatial First) & 92.46\% & 92.90\% & 93.11\% & 93.55\% \\
VideoMamba - M (Temporal First) & 91.8\% & 92.79\% & 92.24\% & 93.01\% \\
\textbf{FuseMamba-VD (Ours)} & 94.64\% & 95.01\% & 95.19\% & \textbf{95.85\%} \\
\hline
\end{tabular}
}
\vspace{-7pt}
\label{tab:cropping_ablation}
\end{table}

We ablate the main components of our architecture on the combined dataset, as it offers a more balanced class distribution and multiple sources/domains, helping to reduce dataset-specific bias and focus on model design. Tab.~\ref{tab:cropping_ablation} shows that all variants benefit from the cropping module and residual skip connections, which consistently improve accuracy by focusing the model on regions where people are likely to appear and stabilize optimization; nevertheless, even with these components, single-branch VideoMamba (including the Large variant with comparable capacity) underperforms our Dual Branch design, indicating that combining spatial- and temporal-first scanning is crucial. In Tab.~\ref{tab:fusion_mechanisms}, we compare fusion mechanisms for lateral connections (LCs): concatenating full hidden states severely degrades performance, likely due to token-space misalignment, so we restrict fusion to CLS tokens, which yields substantial gains; among CLS-based schemes, gated lateral connections from the spatial-first branch (Branch-1) to the temporal-first branch (Branch-2) achieve the best accuracy (95.85\%), outperforming additive, concatenation, and cross-attention fusion. Finally, Tab.~\ref{tab:ablation_lateral} examines where to place lateral connections: while early-, middle-, late-only, and alternating (even/odd) configurations all exceed 94\% accuracy, applying lateral connections continuously at every block yields the highest performance (95.85\%), highlighting the benefit of gradually and repeatedly exchanging information between the two branches throughout the network.

\begin{table}[!h]
\centering
\caption{Performance comparison of different fusion mechanisms.}
\vspace{-4pt}
\scalebox{0.9}{
\renewcommand{\arraystretch}{1.0}
\setlength{\tabcolsep}{6pt}
\begin{tabular}{l|c}
\hline
\textbf{Fusion Mechanism} & \textbf{Accuracy (\%)} \\
\hline
Concatenated LCs (Full Hidden State) & 76.17 \\
Concatenated LCs (CLS Token Only) & 94.64 \\
Additive LCs & 93.44 \\
Cross Attention based LCs & 95.19 \\
Gated LCs (Branch-2 $\rightarrow$ Branch-1) & 93.22 \\
\textbf{Gated LCs (Branch-1 $\rightarrow$ Branch-2)} & \textbf{95.85} \\
\hline
\end{tabular}}
\vspace{-12pt}
\label{tab:fusion_mechanisms}
\end{table}

\begin{table}[!h]
\centering
\caption{Performance comparison for different configurations of lateral connections.}
\vspace{-4pt}
\scalebox{0.9}{
\begin{tabular}{p{6cm}|c}
\hline
\textbf{Lateral-Connection (LC) Config.} & \textbf{Accuracy (\%)} \\
\hline
LCs alternatively (even layers)          & 95.63 \\
LCs alternatively (odd layers)           & 95.63 \\
One LC only at beginning                 & 94.00 \\
One LC only at end                       & 94.43 \\
One LC at middle                         & 94.32 \\
Two LCs at beginning and end             & 95.30 \\
\textbf{Continuous LCs}                  & \textbf{95.85} \\
\hline
\end{tabular}}
\vspace{-12pt}
\label{tab:ablation_lateral}
\end{table}

\section{Conclusion}
\label{sec:conclusion}

In this paper, we present FuseMamba-VD, a novel Dual Branch VideoMamba architecture with Gated Class Token Fusion for violence detection in videos. Our method integrates an efficient state-space model with a dual-stream design aimed at separately capturing fine-grained spatial features and global temporal dynamics. By continuously fusing the spatial class tokens into the temporal branch using a learnable gating mechanism, our approach effectively combines complementary cues, yielding state-of-the-art performance. Extensive experiments on the combined benchmark dataset and the DVD dataset, demonstrate that our model not only achieves state-of-the-art accuracies but also shows significant reductions in model parameters and computational cost in FLOPs compared to previous state-of-the-art methods such as CUE-Net and VideoMamba variants. This work highlights the promise of state-space models for scalable \& resource-efficient violence detection, paving the way for future multi-modal integration.

\textbf{Acknowledgment.} This work was funded through a UKRI EPSRC DTP studentship at Queen Mary University of London. This work utilized Queen Mary's Andrena HPC facility (QMUL Research-IT Services) and resources from the Isambard-AI National AI Research Resource (AIRR), operated by the University of Bristol.

\bibliographystyle{IEEEbib}
\bibliography{icme2026references}

\clearpage
\setcounter{page}{1}
\section{Supplementary Material}
\label{sec:supp}

\subsection{Cropping Module - Supplement}
\label{sec:crop}

\begin{figure}[H]
  \centering
   \includegraphics[width=1\linewidth]{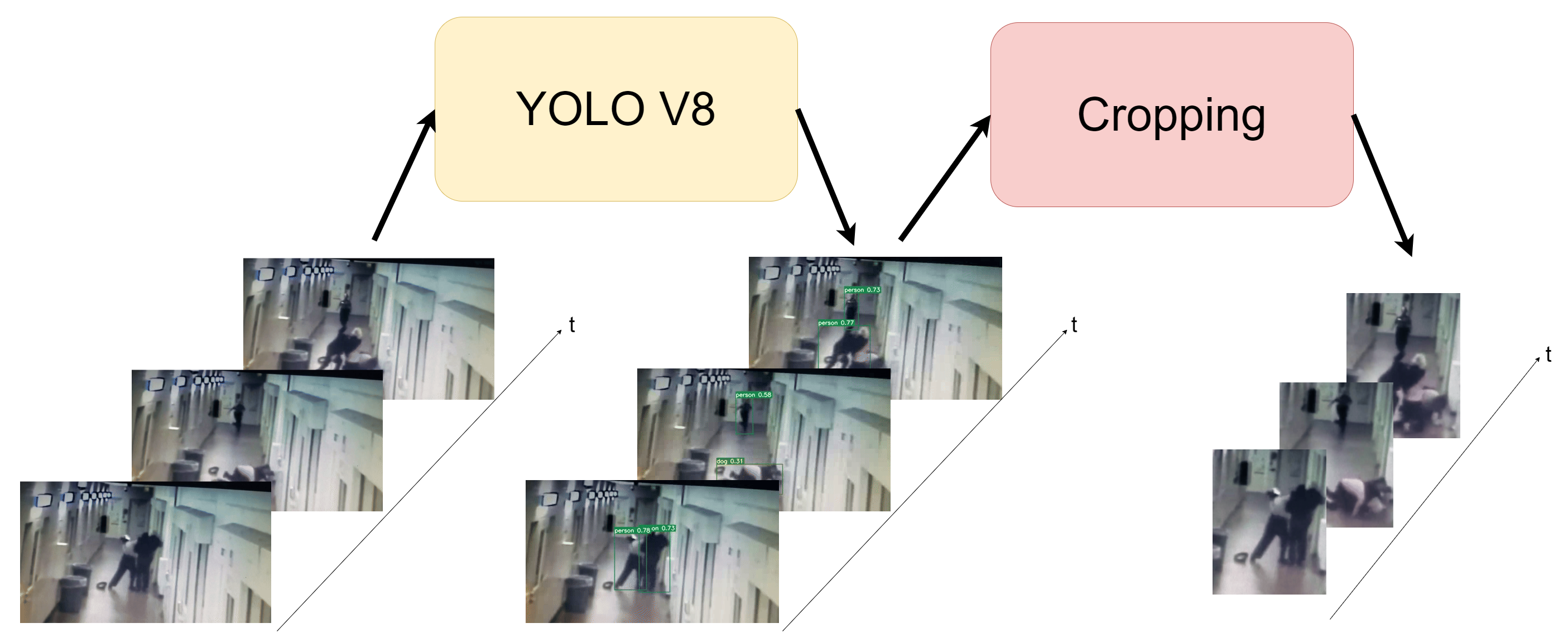}
   \caption{The cropping module which includes YOLO V8 algorithm to detect the people present and crop the maximum bounding region.}
  \label{fig:6}
\end{figure}

The cropping mechanism in the cropping module extracts the region that encompasses all detected people in each frame, to focus specifically on the violent actions taking place, based on the observation that violent incidents typically involve multiple humans. As seen in Fig.~\ref{fig:6}, by computing the maximum bounding area around all detected people, the network is guided to the most informative spatial areas while retaining the original frames if no individuals are detected. Temporal cropping is also not applied here to prevent further potential information loss from missing undetected individuals. YOLO (You Only Look Once) V8 - medium algorithm which classifies objects in a single pass using a CNN-based architecture was used to detect people in this cropping module.

\subsection{Individual DataSet Descriptions - Supplement}

\subsubsection{Real World Fighting (RWF-2000) Dataset}

The Real World Fighting (RWF-2000) dataset was introduced in 2020 and is a comprehensive dataset that contains real world fighting scenarios sourced purely through surveillance footage. This dataset contains 2,000 trimmed video clips and each video is trimmed into 5 seconds where the fighting occurs. The dataset is  balanced, with a 80\%/20\% train-test split which has been thoroughly checked for data leakage between the splits.

\subsubsection{Real Life Violence Situations (RLVS) Dataset}

The Real Life Violence Situations (RLVS) dataset consists of 2000 video clips with 1000 violent and another 1000 non-violent videos collected from YouTube. These contain many real street fight situations in several environments with an average length of 5s from different sources such as surveillance cameras, movies, video recordings, etc. Similar to the RWF-2000 dataset, a 80\%/20\% train-test split has been created for this dataset as well.

\subsubsection{Vision-based Fight Detection From Surveillance Cameras (SURV) Dataset}

SURV is a publicly available dataset created specifically for real-world fight detection from surveillance footage. The videos were collected from YouTube surveillance camera sources and contain fight and non-fight scenes captured in unconstrained environments such as streets, public areas, and institutions. Each video is trimmed into short clips ranging from 2s - 3s and a 80\%/20\% train-test split has been created for this dataset as well. Unlike staged datasets (e.g., movies, hockey), SURV reflects authentic surveillance conditions, including challenges like occlusion, varying viewpoints, and low resolution

\subsubsection{VioPeru Dataset}

In VioPeru dataset, the researchers have created a new balanced dataset called VioPeru, which consists of 280 videos collected from real video surveillance camera records containing challenging violent incidents involving two or more people. The videos have been collected from the citizen security offices of different municipalities in Peru. The videos have been trimmed to 5s just to include the violent incident. Similar to the above datasets, a 80\%/20\% train-test split has also been created.\\

\subsection{Construction of the Combined RWF-2000 + RLVS + SURV + VioPeru Dataset - Supplement}

\paragraph{Dataset selection.}
We consider the four most challenging strongly labeled real-world datasets for violence detection:
RWF-2000, RLVS, SURV~\cite{surv_dataset}, and VioPeru. RWF-2000, SURV, and VioPeru are composed purely of surveillance footage, while RLVS also includes unconstrained online videos. All clips are short, trimmed around the violent event (typically 2--5\,s).

We intentionally exclude:
\begin{itemize}
    \item \textbf{Staged / acted datasets:} AUTO, Movie Scene Violence, and Hockey Fight, which are dominated by cinematic or sports footage.
    \item \textbf{Weakly labeled datasets:} UCF-Crime and XD-Violence, whose video-level labels (Violent / Non-Violent) lack temporal localization and cannot be mixed with strongly labeled data without additional annotation.
\end{itemize}

\paragraph{Duplicate detection and data leakage.}
Although each dataset ships with its own predefined train/test split, directly merging them can introduce cross-dataset leakage, especially since RWF-2000 and RLVS are both partly sourced from YouTube. To detect near-duplicates:
\begin{enumerate}
    \item We extract VideoMAE embeddings for every clip in the training and testing splits of all four datasets.
    \item We compute cosine similarity between every train–test pair.
    \item Pairs with cosine similarity $\geq 0.75$ are manually inspected.
\end{enumerate}
This procedure reveals one duplicated video in the RLVS test split, which we remove. Duplicates within the training sets are retained, as they realistically reflect the frequency of common real-world events, and regularization during training mitigates overfitting to repeated patterns.

\paragraph{Final combined statistics.}
Tab.~\ref{tab:combined_dataset_supp} reports the final per-dataset counts after removing the RLVS test duplicate.

\begin{table}[h]
    \centering
    \caption{Combined RWF-2000 + RLVS + SURV + VioPeru dataset after duplicate removal.}
    \scalebox{0.9}{
    \begin{tabular}{c|cc|cc}
    \hline
         & \multicolumn{2}{c|}{\textbf{Training Set}} & \multicolumn{2}{c}{\textbf{Testing Set}} \\
         & \textbf{Violent} & \textbf{Non-Violent} & \textbf{Violent} & \textbf{Non-Violent} \\ \hline
         \textbf{RWF-2000} & 800 & 800 & 200 & 200 \\
         \textbf{RLVS}     & 800 & 800 & 199 & 200 \\
         \textbf{SURV}     & 120 & 120 & 30 & 30 \\
         \textbf{VioPeru}  & 112 & 112 & 28 & 28 \\ \hline
         \textbf{Total}    & 1832 & 1832 & 457 & 458 \\
    \hline
    \end{tabular}
    }
    \label{tab:combined_dataset_supp}
\end{table}

\subsubsection{Data Leakage Findings after Amalgamation:}

\textbf{SURV vs.\ RWF-2000 and RLVS:} We find no similar videos between the SURV test set and either RWF-2000 or RLVS training sets despite SURV dataset being collected from publicly published surveillance videos in social media.

\textbf{VioPeru vs.\ RWF-2000 and RLVS:} We find no similar videos between the VioPeru test set and either RWF-2000 or RLVS training sets. This is not unexpected, as VioPeru is a newly collected dataset from Peruvian municipalities and contains unique CCTV footage which has not been released on YouTube or any other social media platform.

\textbf{RWF-2000 and RLVS:} We discover one instance where a video in the testing set of RLVS is identical to a video in the training set of RWF-2000. To prevent leakage, we remove the duplicate entry from the RLVS testing set.

\subsection{DVD Dataset - Supplement}
\label{sec:dvd_add}

In our DVD pipeline we did not use fixed windows; instead, we directly segmented from the frame\mbox{-}level annotations by scanning each video’s label stream (aligned via FPS to timestamps) for maximal contiguous runs of the same class and turning those runs into clips: runs labeled $0$ were recorded as \textit{violent} clips and runs labeled $1$ as \textit{non\mbox{-}violent} clips, while any frames labeled $-1$ (irrelevant/uncertain) were treated as “ignore.” Whenever a $-1$ label occurred, the corresponding portion was discarded rather than assigned to either class, and any runs bordering a $-1$ region were truncated at the boundary so no ambiguous frames leaked into the clips. For each retained run we stored start and end times, duration, FPS, width/height, the source file name, and a stable relative path in a manifest, ensuring clips never overlap across classes and that every exported segment is a faithful, contiguous slice of unambiguous labels derived directly from the annotations. Media extraction was performed only after the manifest was finalized (late binding), snapping cut points to sensible decoding boundaries; no gap\mbox{-}merging or window resampling was applied, and no re\mbox{-}labeling heuristics were used—clip identities come solely from contiguous $0/1$ runs with all $-1$ spans excluded.

On the resulting clip set, we have 2,648 clips in total ($\approx 24.60$ hours of video). Of these, 1,099 violent clips (41.50\%) contribute $574.53$ minutes, and 1,549 non-violent clips (58.50\%) contribute $901.53$ minutes. Clip durations are heterogeneous by design: violent clips have a median $14.02\,\mathrm{s}$ (mean $31.37\,\mathrm{s}$; IQR $5.04$--$37.04\,\mathrm{s}$; min $1.00\,\mathrm{s}$; max $575.04\,\mathrm{s}$), while non-violent clips have a median $8.94\,\mathrm{s}$ (mean $34.92\,\mathrm{s}$; IQR $1.97$--$26.00\,\mathrm{s}$; min $0.03\,\mathrm{s}$; max $1{,}713.28\,\mathrm{s}$). This skew, especially the long tail for non-violent background, reflects real footage composition and helps models learn context without leaking into violent transitions. The videos span a wide range of capture settings: FPS from $13.14$ to $60.00$ (most common: $\sim29.97$, $25.00$, $30.00$), and resolutions from $192\times240$ up to $3840\times3840$, with $1920\times1080$ most frequent (followed by $3840\times2160$ and $1280\times720$).

\subsubsection{DVD Clip Segmentation and Train/Test Split Protocol}

\paragraph{Frame-level labels and clip segmentation.}
The DVD dataset provides frame-level annotations (violent vs.\ non-violent) over 344 long videos (2.7M frames). To convert these into clip-level samples:
\begin{enumerate}
    \item We scan each video’s frame-level labels.
    \item Maximal contiguous runs of violent frames are grouped into a single violent clip.
    \item Maximal contiguous runs of non-violent frames are grouped into a single non-violent clip.
\end{enumerate}
This yields 2{,}648 clips: 1{,}099 violent and 1{,}549 non-violent, corresponding to approximately 9.5 and 15 hours of footage, respectively (24.6 hours total).

\paragraph{Source-level train/test partitioning.}
To prevent data leakage, we enforce a strict source-level split:
\begin{itemize}
    \item All clips originating from a given source video are assigned either to the training set or to the test set, never both.
    \item We select an approximately 80/20 split at the video level, while preserving the overall violent/non-violent ratio.
\end{itemize}

The resulting statistics are summarized in Tab.~\ref{tab:dvd_split_stats_supp}.

\begin{table}[h]
\centering
\caption{DVD train/test split used in our experiments (clip-level view).}
\scalebox{0.95}{
\begin{tabular}{l|c|c|c}
\hline
 & \textbf{Violent} & \textbf{Non-Violent} & \textbf{Total} \\
\hline
\textbf{Training Set} & 820 & 1167 & 1987 \\
\textbf{Testing Set}  & 279 & 382  & 661  \\
\hline
\textbf{Total}        & 1099 & 1549 & 2648 \\
\hline
\end{tabular}}
\label{tab:dvd_split_stats_supp}
\end{table}

\subsubsection{Per-dataset generalizability - FuseMamba-VD Performance on separate datasets of RWF-2000, RLVS SURV, and VioPeru - Supplement}
\label{sec:sep_perform}

When tested separately, our architecture outperforms previously reported state-of-the-art classification accuracies in the literature for RWF-2000, RLVS and SURV datasets, by achieving accuracies of 94.50\%, 99.75\% and 96.67\% respectively, setting a new state-of-the-art. For the VioPeru dataset, our FuseMamba-VD architecture is able to reach the already reported state-of-the-art accuracy of 89.23\%.

\begin{table}[!h]
\centering
\caption{Comparison of earlier best accuracies with FuseMamba-VD on four datasets.}
\begin{tabular}{l|c|c}
\toprule
\textbf{Dataset} & \makecell*[c]{\textbf{Reported Best}\\\textbf{Accuracy (\%)}} & \makecell*[c]{\textbf{Our Model}\\\textbf{Accuracy (\%)}} \\
\midrule
RWF-2000 & 94.36    & \textbf{94.50}  \\
VioPeru  & \textbf{89.23} & \textbf{89.23} \\
RLVS     & 99.50~  & \textbf{99.75} \\
SURV     & 95.62  & \textbf{96.67} \\
\bottomrule
\end{tabular}
\label{tab:individual_datasets}
\end{table}

\subsection{Implementation Details - Supplement}
\label{sec:imp_details_sup}

Our FuseMamba-VD architecture was implemented in PyTorch using the AdamW optimizer with a cosine learning rate schedule starting with a learning rate of 1.5e-5 and Cross-Entropy Loss, taking insights from training recipes of the original VideoMamba architecture. Each branch was initialized with pre-trained weights of Vanilla VideoMamba trained with Kinetics-400. All models were trained for 55 epochs with 5 warm-up epochs where the best validation model was saved. The final training hyperparameters can be found in Tab.~\ref{tab:hyper_parameters}.

\begin{table}[tp]
	\centering
    \caption{
        Training hyperparameters for our experiments.}
    \resizebox{0.5\textwidth}{!}{
      	\begin{tabular}{lc}
    	    \Xhline{1.0pt}
    	    ~ & Combined Dataset/ DVD Dataset  \\
    	    \Xhline{0.8pt}
    	    \textit{Optimization} & ~ \\
    	    Optimizer & AdamW \\
    	    Momentum & $\beta_1, \beta_2=0.9, 0.999$ \\
    	    Weight decay & 0.05 \\
    	    Learning rate schedule & cosine decay \\
    	    Batch size & 16 \\
    	    Learning rate & 1.5e-5 \\
    	    Warmup epochs & 5 \\
    	    Total epochs & 55 \\
    	    \hline
    	    \textit{Data augmentation} & ~ \\
    	    Inception-style cropping & ~ \\
    	    \ \ \ \ Scale & [0.08, 1.00] \\
    	    \ \ \ \ Jitter aspect ratio & [0.75, 1.33] \\
    	    Color jitter & 0.4 \\
    	    Rand augment & rand-m7-n4-mstd0.5-inc1 \\
    	    Repeated sampling & 1 \\
    	    \hline
    	    \textit{Regularisation} & ~ \\
    	    Dropout & 0.5 \\
    	   
    	    \bottomrule
    	\end{tabular}
    }
    \label{tab:hyper_parameters}

\end{table}

\begin{table}[!h]
  \centering
  \caption{Results comparison for the RLVS Dataset.}
  \begin{tabular}{ p{2.25cm} p{2.5cm} p{2cm}}
    \toprule
    \makecell*[c]{Method} & \makecell*[c]{Model Type} & \makecell*[c]{Accuracy (\%)}\\
    \midrule
    \makecell*[c]{CNN-\\LSTM} & \makecell*[c]{VGG16+LSTM} & \makecell*[c]{88.20} \\
    
    \makecell*[c]{Temporal \\ Fusion CNN \\+LSTM} & \makecell*[c]{CNN+LSTM} & \makecell*[c]{91.02} \\
    
    \makecell*[c]{DeVTr} & \makecell*[c]{ViViT} & \makecell*[c]{96.25} \\

    \makecell*[c]{ACTION-\\VST} & \makecell*[c]{CNN + ViViT} & \makecell*[c]{98.69}\\

     \makecell*[c]{CUE-Net} & \makecell*[c]{Enhanced \\UniformerV2 \\}  & \makecell*[c]{99.50}\\

     \makecell*[c]{Video- \\ Mamba} & \makecell*[c]{\\ SSM \\ }  & \makecell*[c]{99.50}\\

    \makecell*[c]{\textbf{Fuse} \\ \textbf{Mamba-VD}} & \makecell*[c]{\\ \textbf{SSM} \\ }  & \makecell*[c]{\textbf{99.75}}\\
   
    \bottomrule
  \end{tabular}
  \label{tab:rlvs}
\end{table}

\begin{table}[!h]
  \centering
  \caption{Results comparison for the RWF-2000 Dataset.}
  \begin{tabular}{ p{2.25cm} p{2.5cm} p{2cm}}
    \toprule
    \makecell*[c]{Method} & \makecell*[c]{Model Type} & \makecell*[c]{Accuracy (\%)}\\
    \midrule
    \makecell*[c]{ConvLSTM} & \makecell*[c]{CNN+LSTM} & \makecell*[c]{77.00} \\
    \makecell*[c]{X3D} & \makecell*[c]{3DCNN} & \makecell*[c]{84.75} \\
    \makecell*[c]{I3D} & \makecell*[c]{3DCNN} & \makecell*[c]{83.40} \\
    \makecell*[c]{Flow Gated \\ Network} & \makecell*[c]{Two Stream \\ Graph CNN} & \makecell*[c]{87.25} \\
    \makecell*[c]{SPIL} & \makecell*[c]{Graph CNN} & \makecell*[c]{89.30} \\
   \makecell*[c]{Structured \\ Keypoint \\ Pooling} & \makecell*[c]{CNN} & \makecell*[c]{93.40} \\
   \makecell*[c]{Video Swin \\ Transfor-\\mer} & \makecell*[c]{ViViT} & \makecell*[c]{91.25} \\
   \makecell*[c]{ACTION-\\VST} & \makecell*[c]{CNN + ViViT} & \makecell*[c]{93.59}\\

    \makecell*[c]{CUE-Net} & \makecell*[c]{Enhanced \\UniformerV2 \\}  & \makecell*[c]{94.00}\\

    \makecell*[c]{Video- \\ Mamba} & \makecell*[c]{\\ SSM \\ }  & \makecell*[c]{92.75}\\

      \makecell*[c]{Multi- \\ Head Att \\ \& LSTM } & \makecell*[c]{LSTM + ViViT}  & \makecell*[c]{94.36}\\

    \makecell*[c]{\textbf{Fuse} \\ \textbf{Mamba-VD} } & \makecell*[c]{\\ \textbf{SSM} \\ }  & \makecell*[c]{\textbf{94.50}}\\
    \bottomrule
  \end{tabular}
  \label{tab:rwf}
\end{table}

\begin{table}[!h]
  \centering
  \caption{Results comparison for the VioPeru Dataset.}
  \begin{tabular}{ p{2.25cm} p{2.5cm} p{2cm}}
    \toprule
    \makecell*[c]{Method} & \makecell*[c]{Model Type} & \makecell*[c]{Accuracy (\%)}\\
    \midrule

    \makecell*[c]{Sep \\ Conv  \\ LSTM-} & \makecell*[c]{CNN + LSTM} & \makecell*[c]{73.21}\\

    \makecell*[c]{\textbf{Advanced}\\ \textbf{Sep} \\ \textbf{Conv}  \\ \textbf{LSTM}} & \makecell*[c]{\textbf{CNN + LSTM}} & \makecell*[c]{\textbf{89.23}}\\

     \makecell*[c]{Video- \\ Mamba} & \makecell*[c]{\\ SSM \\ }  & \makecell*[c]{85.71}\\

    \makecell*[c]{\textbf{Fuse} \\ \textbf{Mamba-VD} } & \makecell*[c]{\\ \textbf{SSM} \\ }  & \makecell*[c]{\textbf{89.23}}\\
   
    \bottomrule
  \end{tabular}
  \label{tab:vperu}
\end{table}

\begin{table}[!h]
  \centering
  \caption{Results comparison for the SURV Dataset.}
  \begin{tabular}{ p{2.25cm} p{2.5cm} p{2cm}}
    \toprule
    \makecell*[c]{Method} & \makecell*[c]{Model Type} & \makecell*[c]{Accuracy (\%)}\\
    \midrule

    \makecell*[c]{Temporal \\ Spatial  \\ Attn. Maps} & \makecell*[c]{CNN + Attn} & \makecell*[c]{91.80}\\

    \makecell*[c]{RTFM } & \makecell*[c]{MIL based} & \makecell*[c]{95.62}\\

     \makecell*[c]{Video- \\ Mamba} & \makecell*[c]{\\ SSM \\ }  & \makecell*[c]{95.00}\\

    \makecell*[c]{\textbf{Fuse} \\ \textbf{Mamba-VD}} & \makecell*[c]{\\ \textbf{SSM} \\ }  & \makecell*[c]{\textbf{96.67}}\\
   
    \bottomrule
  \end{tabular}
  \label{tab:surv}
\end{table}

\subsection{Statistical Significance Analysis - Supplement}
\label{sec:statsig}

We compare FuseMamba-VD with VideoMamba-Large using McNemar’s two-sided exact test.

\paragraph{Combined dataset.}
The contingency table between the two models yields
$n_{01} = 15$, $n_{10} = 46$, $d = n_{01} + n_{10} = 61$, and
$p = 8.84 \times 10^{-5}$, indicating a statistically significant improvement at $p < 0.05$.

\paragraph{DVD dataset.}
For DVD, we obtain
$n_{01} = 32$, $n_{10} = 52$, $d = 84$, and $p = 0.0375$,
again showing a statistically significant difference at $p < 0.05$.

These results support the conclusion that FuseMamba-VD’s gains over VideoMamba-Large are not due to chance.

We report McNemar’s test against VideoMamba-Large only (but not against any other model) because it is the closest controlled baseline to FuseMamba-VD, sharing the same SSM family, Kinetics-400 pretraining, and nearly matched model capacity. This makes the paired comparison methodologically more rigorous for isolating the contribution of our dual-branch fusion design, whereas for example CUE-Net is treated primarily as a heterogeneous accuracy/efficiency reference.

\subsection{Visualization of Spatial-First and Temporal-First Branches - Supplement}
\label{sec:supp_cam}

We employ Grad-CAM to generate class activation maps (CAMs) that visualize the regions influencing the model’s predictions. In the spatial-first branch, this approach is particularly effective because the Mamba blocks primarily extract spatial features and object-level cues, and the resulting gradients highlight discriminative image regions. The temporal-first branch, in contrast, is designed to prioritize motion dynamics over high-resolution spatial structure. As a result, deeper layers in this branch tend to aggregate information over time, causing their spatial representations to become highly compressed and globally distributed. Conventional Grad-CAM often fails in this setting, as the gradients of late features are dominated by temporal aggregation, leading to diffused or uninformative maps.

To address this limitation in the temporal-first branch, we employ Grad-CAM++, which introduces higher-order gradient weighting and explicitly accounts for the contribution of multiple activation regions. Formally, the Grad-CAM++ coefficients $\alpha_{ij}^{kc}$ associated with spatial location $(i,j)$ of feature map $A^k$ for class $c$ are given by:
\begin{equation}
\alpha_{ij}^{kc}
=
\frac{
\frac{\partial^2 y^c}{\partial (A_{ij}^k)^2}
}{
2 \frac{\partial^2 y^c}{\partial (A_{ij}^k)^2}
+
\sum_{p,q} A_{pq}^k
\frac{\partial^3 y^c}{\partial (A_{ij}^k)^3}
},
\end{equation}
where $y^c$ denotes the logit for class $c$. These coefficients determine how strongly each spatial location contributes to the class score. Unlike standard Grad-CAM, which effectively assumes a single dominant activation region, Grad-CAM++ can naturally model multiple disjoint or evolving regions of interest, which is characteristic of violent interactions unfolding over time. Because the temporal-first branch encodes temporal patterns densely, the gradients in this branch highlight all spatial locations where motion cues contributed to the predicted violence class, rather than a single frame or region. This produces spatially spread heatmaps in which the highlighted regions align with the trajectory of the violent motion (e.g., swinging arms, body movements, approach/retreat dynamics). In other words, the temporal-first pathway compresses spatial detail but enriches temporal cues, so evidence accumulates across many pixels over time; Grad-CAM++ might be interpreting these dispersed temporal contributions by assigning relevance to multiple active locations simultaneously. It indicates that the temporal-first branch is not relying on a single static pose but instead on motion-consistent evidence across frames.

\begin{figure}
  \centering
   \includegraphics[width=1\linewidth]{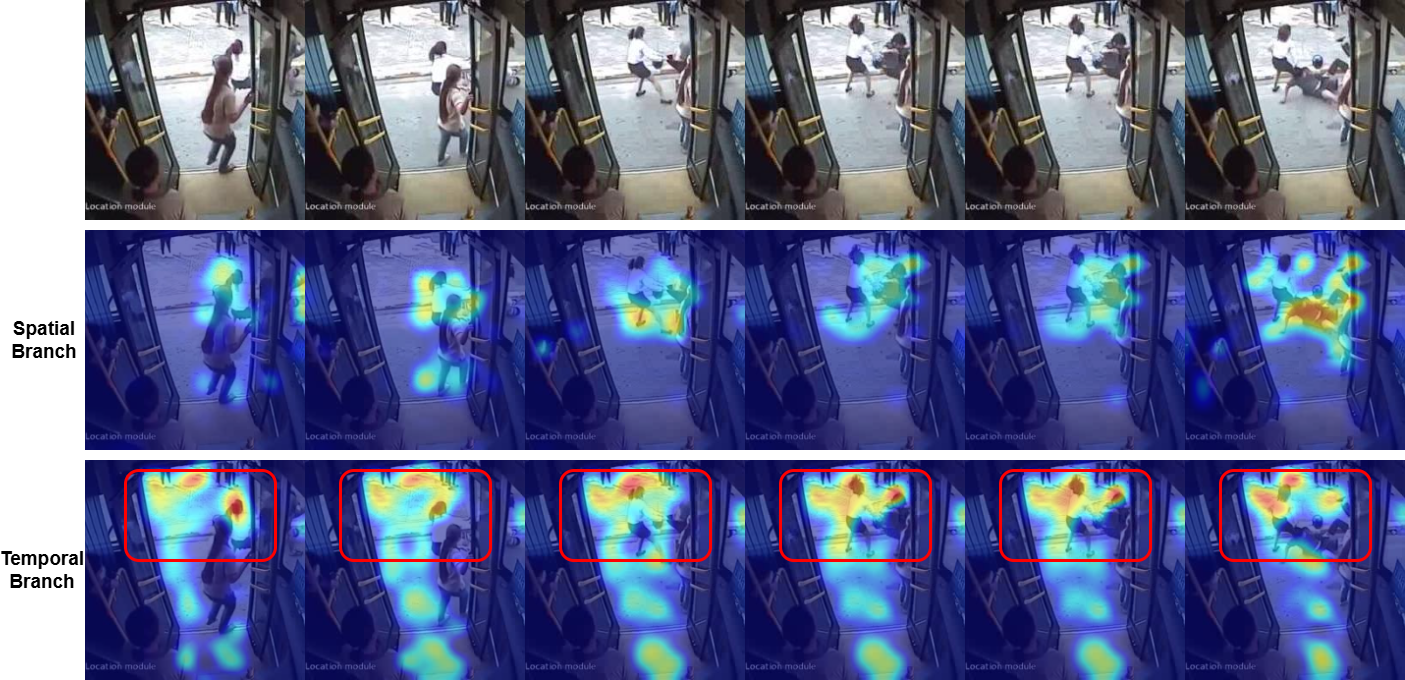}

   \caption{Visual Analysis of Class Activation Maps (CAMs). CAMs for another violent-labeled video correctly classified as violent.}

  \label{fig:4}
\end{figure}

Accordingly, in our analysis we use Grad-CAM on the spatial-first branch to obtain fine-grained spatial CAMs, while Grad-CAM++ is applied to the temporal-first branch to interpret motion-related contributions. Fig.~\ref{fig:3} from the main paper and Fig.~\ref{fig:4} from this supplementary section illustrates representative violent video examples where the key violent action is partially occluded within the scene. The spatial-first branch successfully localizes the relevant interaction despite this occlusion, as evidenced by the concentrated Grad-CAM heatmap highlighting the precise region of the violent act. In contrast, the temporal-first branch produces a more distributed activation pattern that captures the evolving trajectory of the incident across frames, resulting in a spatially extended hotspot following the motion of the aggressor and victim over time. This complementary behavior demonstrates that while the spatial branch excels at pinpointing the exact location of the partially hidden action, the temporal branch effectively traces its motion dynamics and temporal extent. To further verify that this pattern is not a ghosting artifact of our frame sampling strategy, we also visualized Grad-CAM++ on inputs consisting of 64 strictly consecutive frames and observed the same localized and trajectory-aware activation patterns for a sliding window, confirming that these maps reflect genuine model behavior rather than temporal leakage.

\subsection{Failure Case Analysis - Supplement}
\label{sec:faliure_case_supsep_perform}

\begin{figure}
    \centering
    \includegraphics[width=1\linewidth]{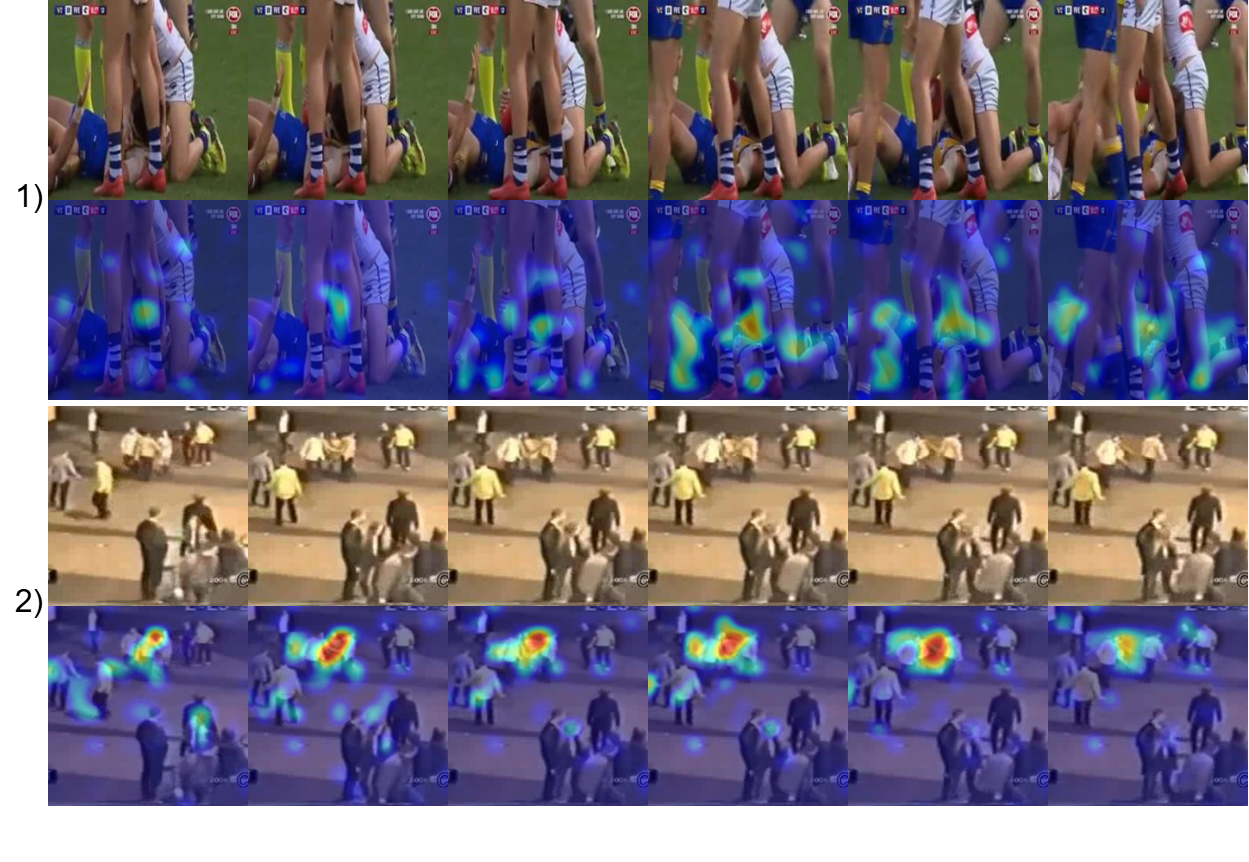}
    \caption{ Failure Case Analysis.}
    \label{fig:7}
\end{figure}

Fig.~\ref{fig:7} shows the Grad-CAMs from the spatial-first branch for two non-violent videos that the model mistakenly classified as violent. In the first example, the model highlights the area where a rugby ball is being forcefully removed, suggesting it may perceive such forceful actions as violent. In the second example, despite several interactive groups being present, the CAMs concentrate on a collision involving a woman and a group, which might have been a violent scenario in reality. These observations indicate that, while the model has effectively learned to pinpoint cues related to violent actions, it may sometimes overemphasize certain visual signals as well.

\subsection{Ablation Study - Supplement}
\label{sec:abl4_details_sup}

\subsubsection{Ablation on Number of Frames for each Branch}

Tab.~\ref{tab:ablation_number_frames} examines the impact of varying the number of frames inputted to each branch. This ablation study shows that providing each branch with the same frame count appears beneficial when it comes to our FuseMamba-VD architecture, likely due to more balanced representation learning and consistent temporal context across branches.

\begin{table}[!h]
\centering
\caption{Comparison of model performance with different number of frames inputted into 2 branches evaluated against the combined dataset.}
\scalebox{1}{
\begin{tabular}{c|c|c}
\toprule
\textbf{Branch-1} & \textbf{Branch-2} & \textbf{Accuracy (\%)} \\
\midrule
32 & 64 & 94.75 \\
64 & 32 & 95.08 \\
32 & 32 & 95.74 \\
\textbf{64} & \textbf{64} & \textbf{95.85} \\
\bottomrule
\end{tabular}
}
\label{tab:ablation_number_frames}
\end{table}

\subsubsection{Ablation on Fusion Mechanism in Final Fusion Block}
\label{sec:abl_final_fusion}

Tab.~\ref{tab:fusion_mechanisms_final} compares several ways of merging the two branch outputs in the final fusion block on the combined dataset. Overall, the gap between methods is small, with simple concatenation giving the best accuracy 95.85\%, narrowly ahead of cross-attention 95.74\% and the addition baseline 94.97\%. We adopt concatenation due to its simplicity and parameter efficiency. It matches and slightly surpasses cross-attention while avoiding the extra query–key value projections and attention maps, thereby reducing compute, memory, and latency. The gated lateral connections (LCs) seem more sensitive to direction. Propagating information from Branch-1$\rightarrow$Branch-2 is competitive (95.30\%), whereas the reverse Branch-2$\rightarrow$Branch-1 degrades the accuracy (93.77\%). Given its robustness, simplicity, and lower complexity, we chose to use concatenation in the final fusion block.

\begin{table}[!h]
\centering
\caption{Performance comparison of different fusion mechanisms in the final fusion block for the combined dataset.}
\scalebox{0.92}{
\renewcommand{\arraystretch}{1.0}
\setlength{\tabcolsep}{6pt}
\begin{tabular}{l|c}
\hline
\textbf{Fusion Mechanism} & \textbf{Accuracy (\%)} \\
\hline
Addition & 94.97 \\
Cross Attention & 95.74 \\
Gated LCs (Branch-2 $\rightarrow$ Branch-1) & 93.77 \\
Gated LCs (Branch-1 $\rightarrow$ Branch-2) & 95.30 \\
\textbf{Concatenation} & \textbf{95.85} \\
\hline
\end{tabular}}
\label{tab:fusion_mechanisms_final}
\end{table}

\end{document}